\documentclass[journal]{IEEEtai}

\usepackage[colorlinks,urlcolor=blue,linkcolor=blue,citecolor=blue]{hyperref}

\usepackage{color,array}

\usepackage{graphicx}

\usepackage{cite}
\usepackage{amsmath,amssymb,amsfonts}
\usepackage{amsmath,amsfonts,bm}
\usepackage{algorithmic}
\usepackage{algorithm,algorithmic}

\usepackage{textcomp}
\usepackage{url}
\usepackage{graphicx}
\usepackage{pgf}
\usepackage[modulo,right]{lineno}
\usepackage{booktabs}
\usepackage{multirow}

\newcommand{\rev}[1]{{\color{black}{#1}}}

\begin{document}

\title{Augmenting generative models with biomedical knowledge graphs improves targeted drug discovery}

\author{Aditya Malusare, Vineet Punyamoorty and Vaneet Aggarwal\thanks{
A. M. and V. A. are with the Edwardson School of Industrial Engineering and the Institute of Cancer Research, Purdue University, email: \{malusare, vaneet\}@purdue.edu. V. P. is with the Elmore Family School of Electrical and Computer Engineering, Purdue University, email: vpunyamo@purdue.edu

A. M. gratefully acknowledges the Walther Cancer Foundation and support from the Purdue University Institute for Cancer Research, P30CA023168. This work was supported in part by the National Science Foundation under grant FW-HTF-R-2129097.

This paper has been accepted for publication in the IEEE Transactions on Artificial Intelligence, October 2025

© 2025 IEEE. Personal use of this material is permitted. Permission from IEEE must be obtained for all other uses, in any current or future media, including reprinting/republishing this material for advertising or promotional purposes, creating new collective works, for resale or redistribution to servers or lists, or reuse of any copyrighted component of this work in other works.
}}

\markboth{IEEE Transactions on Artificial Intelligence}{A. Malusare, V. Punyamoorty, and V. Aggarwal: Augmenting generative models with biomedical knowledge graphs}

\maketitle

\begin{abstract}
	Recent breakthroughs in generative modeling have demonstrated remarkable capabilities in molecular generation, yet the integration of comprehensive biomedical knowledge into these models has remained an untapped frontier. In this study, we introduce K-DREAM (Knowledge-Driven Embedding-Augmented Model), a novel framework that leverages knowledge graphs to augment diffusion-based generative models for drug discovery. \rev{By embedding structured information from large-scale knowledge graphs, K-DREAM directs molecular generation toward candidates with higher biological relevance and therapeutic suitability. This integration ensures that the generated molecules are aligned with specific therapeutic targets, moving beyond traditional heuristic-driven approaches. In targeted drug design tasks, K-DREAM generates drug candidates with improved binding affinities and predicted efficacy, surpassing current state-of-the-art generative models. It also demonstrates flexibility by producing molecules designed for multiple targets, enabling applications to complex disease mechanisms. These results highlight the utility of knowledge-enhanced generative models in rational drug design and their relevance to practical therapeutic development.}
\end{abstract}
\vspace{-5pt}
\begin{IEEEImpStatement}
We introduce K-DREAM, a new approach to drug discovery that combines knowledge graphs with AI-driven drug design. Unlike conventional methods that focus mainly on chemical properties, our framework incorporates biological relationships to create more medically relevant drug candidates. When tested on various protein targets, our system produced potential drugs with better binding abilities while maintaining diversity and feasibility for manufacturing. Particularly promising is K-DREAM's ability to design drugs that work on multiple targets simultaneously, a crucial advantage for treating complex diseases that involve multiple biological pathways. By bringing biological context directly into the drug design process, our work could significantly reduce the time and resources needed to discover promising new treatments, potentially accelerating the journey from laboratory to patient.
\end{IEEEImpStatement}


\section{Introduction}

\begin{figure*}
\centering
\includegraphics[width=.98\textwidth]{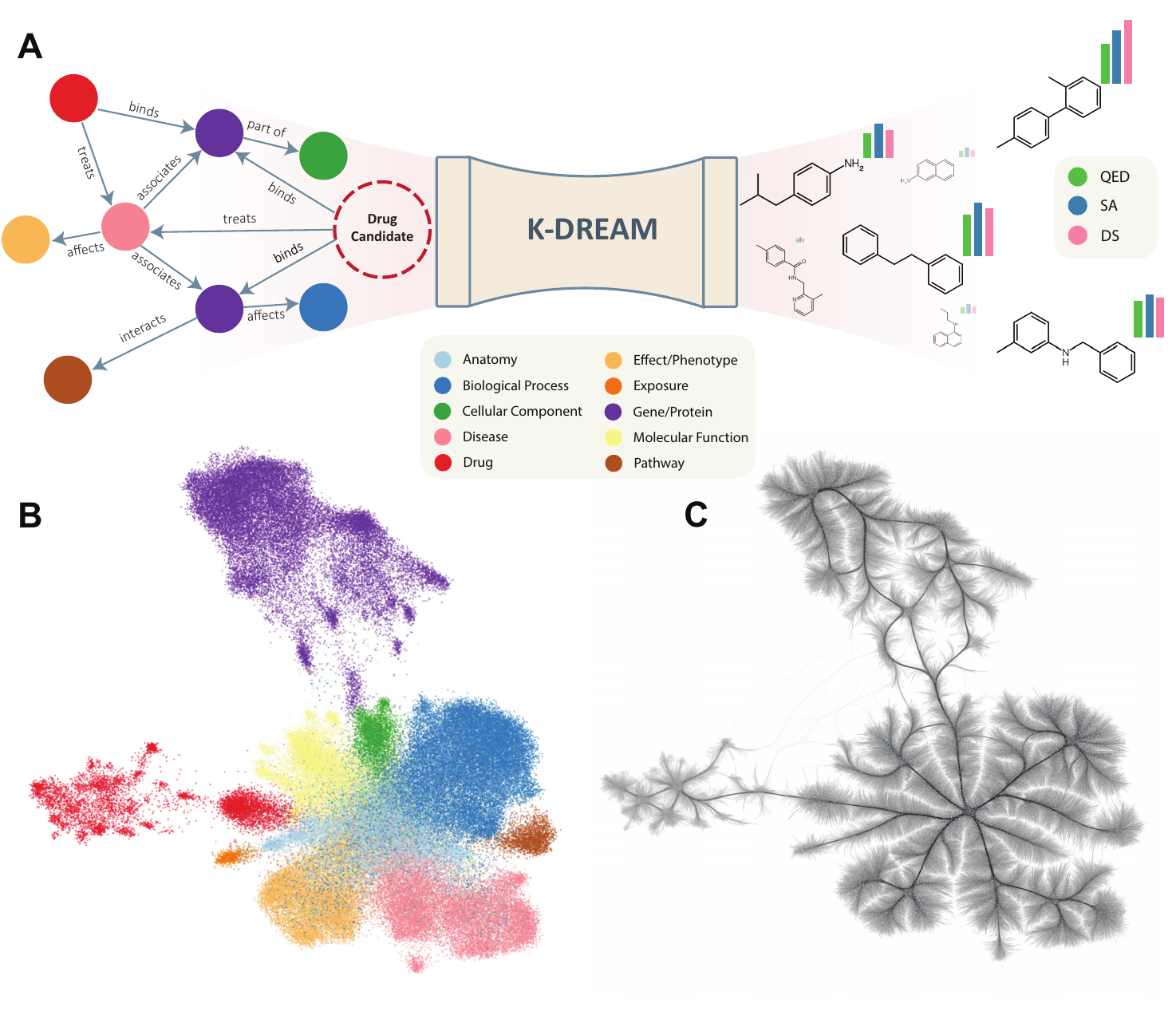}
\caption{ \textbf{The Knowledge-Driven Embedding-Augmented Model (K-DREAM).} \textbf{A} Overview of the K-DREAM generative model for molecular structures, incorporating guidance information from embeddings derived from the knowledge graph. The diffusion process is guided to produce molecules that are both chemically valid and biologically relevant to a given target embedding. Molecules are evaluated using metrics such as docking score, Quantitative Estimate of Drug-likeness (QED), and Synthetic Accessibility (SA). \textbf{B} Embeddings of the PrimeKG \cite{Chandak2023} knowledge graph created using the TransE \cite{NIPS2013_1cecc7a7} model, projected using the Uniform Manifold Approximation Projection (UMAP) algorithm. \textbf{C} Visualization of the relationships between UMAP-projected PrimeKG embeddings using edge bundling (a technique to reduce visual clutter in network visualizations by grouping edges that follow similar paths). Bundled edges represent relationships between entities, with thicker bundles indicating stronger or more numerous connections between related clusters.}
\label{fig:main}
\label{fig}
\end{figure*}

Generative models have achieved significant success in fields such as text and image synthesis, establishing state-of-the-art performance in natural language processing, computer vision, and decision making \cite{cao2024survey,chen2024deep}. These successes have inspired the extension of generative models to other modalities, including graph-based representations of molecular structures, which hold considerable potential in drug discovery. Molecular generative models, in particular, have emerged as powerful tools for de novo drug design and optimization, leveraging graph-based representations to navigate the complex chemical space. Techniques such as variational autoencoders (VAEs) \cite{jin2020hierarchical}, generative adversarial networks (GANs), autoregressive models \cite{kong2023autoregressive}, and reinforcement learning \cite{jeon20morld,olivecrona2017molecular,yang21freed} have been employed to explore molecular configurations and propose novel therapeutic candidates.

Despite their promise, these models are often constrained by their reliance on simplistic guidance mechanisms, typically involving heuristic chemical scores, such as Synthetic Accessibility (SA) \cite{ertl2009estimation} or Quantitative Estimate of Drug-likeness (QED) \cite{bickerton2012quantifying}, along with structural constraints like validity and uniqueness. These conventional approaches largely overlook the extensive body of biomedical knowledge accumulated through years of research and clinical practice, which could significantly enhance the generative process.

Biomedical knowledge is systematically encoded into extensive knowledge graphs \cite{Chandak2023,himmelstein2017systematic}, which span millions of relations and provide structured representations of interactions within biological systems. These graphs capture relationships across multiple biological scales, from individual molecular components, such as genes, proteins, and small molecules, to higher-order entities like cells, tissues, and entire biological processes. The relationships in these graphs reflect diverse biological phenomena, including protein-protein interactions, gene regulation, metabolic pathways, and drug-target associations.

\rev{The structured format of biomedical knowledge graphs captures complex biological behaviors that arise from interactions between molecules. For example, cellular homeostasis (the ability of cells to maintain stable internal conditions) is driven by feedback loops within signaling pathways \cite{harris2005p53}. Similarly, phenotypic robustness, which refers to an organism’s ability to maintain a stable phenotype despite genetic changes or environmental fluctuations, often emerges from redundant genetic networks \cite{wagner2000robustness}. A major challenge in drug development is drug resistance, which can occur through adaptive mechanisms such as the overactivation of efflux pumps (proteins that expel drugs from cells) or mutations in drug target proteins \cite{fernandez2012adaptive}. Biomedical knowledge graphs also support drug repurposing by revealing hidden relationships between diseases and existing drugs \cite{mohamed2020discovering}, as demonstrated by the identification of baricitinib for COVID-19 treatment by linking its known mechanisms with pathways involved in viral infectivity \cite{vaccines10060951}. By capturing these diverse and interconnected biological relationships, knowledge graphs provide a rich source of information for generative models to design drug candidates that are biologically meaningful.} Existing work on integrating knowledge graphs primarily focuses on enhancing molecular property prediction tasks, such as predicting interactions \cite{lin2020kgnn}, repurposing \cite{bang2023biomedical}, and molecular property optimization \cite{zeng2022toward}. However, the potential of knowledge graphs to guide the generative process itself remains largely unexplored.

In this paper, we introduce a novel framework called K-DREAM (Knowledge-Driven Embedding-Augmented Model), which bridges the gap between state-of-the-art generative models and the wealth of information contained within biomedical knowledge graphs (Figure \ref{fig:main}). Our approach aims to generate biologically relevant and therapeutically promising molecules by leveraging biomedical knowledge graphs to inform and guide molecular generation. Specifically, we integrate the embeddings derived from knowledge graphs, preserving the semantic integrity of the biological information, into a generative framework. By combining the power of diffusion-based generative modeling with knowledge graph embeddings, K-DREAM produces novel drug candidates that are not only chemically valid but also enriched with specific therapeutic properties, enhancing their potential in drug development. \rev{ Prior work on applied generative models in this field uses conditional diffusion-based architectures to generate molecular graphs to maximize previously mentioned heuristic scores like QED and SA, with techniques to adapt to optimizing for these non-differentiable metrics.} We demonstrate that the embeddings derived from biomedical knowledge graphs contain sufficient information to steer the generative process towards biologically relevant molecules, thereby improving the quality and relevance of the generated compounds without explicitly including quantitative scores or structural constraints.

To represent the biomedical knowledge in a form compatible with generative models, we use knowledge graph embedding (KGE) techniques, which transform entities and relationships in the graph into a continuous vector space while preserving semantic relationships. In this work, we employ the TransE model \cite{NIPS2013_1cecc7a7} due to its interpretability and efficiency. TransE represents relations as translations between entities in a real-valued embedding space, making it suitable for conceptualizing molecular generation as a path through a unified biological embedding space. This additive nature of TransE allows us to reason intuitively about entity relationships and integrate these insights into the generative modeling process.

K-DREAM outperforms state-of-the-art molecular generative models on targeted drug design tasks, producing compounds with improved biological relevance and therapeutic potential. In docking studies against specific protein targets, K-DREAM-generated molecules consistently achieved higher docking scores compared to other models, indicating greater binding likelihood to intended targets. The framework's adaptability enables diverse generative tasks, including multi-target drug design for compounds with tailored polypharmacological profiles. We demonstrate this by generating molecules designed to interact with multiple protein targets simultaneously, exploring complex drug design scenarios potentially addressing diseases with multiple underlying mechanisms. 

\section{Methods}

K-DREAM systematically bridges molecular generation with biomedical knowledge through four key components. We begin with foundational representations of both molecular structures and biomedical knowledge, establishing the mathematical framework for our approach. The molecular generation pipeline then builds from an unconditional base model to a knowledge-guided framework, incorporating information from biomedical relationships to steer the generative process. We develop specialized neural architectures to enable precise navigation of the resulting chemical-knowledge space, allowing targeted molecular design. Finally, we conclude with techniques for exploration and visualization of this unified space, followed by details of our experimental setup for reproducibility. 

\subsection{Molecular Representation}

Molecular structures are represented in this work using a planar graph $\textbf{G} = (\textbf{X}, \textbf{E})$ where $\textbf{X}\in\mathbb{R}^{N\times M}$ is a feature matrix for $N$ nodes (heavy atoms) described by $M$-dimensional vectors encoding atom information, and $\textbf{E}\in\mathbb{R}^{N\times N}$ is the adjacency matrix indicating the presence of single, double or triple bonds between the nodes. 

\subsection{Knowledge Graph Embeddings}

We generate the knowledge graph embeddings (KGEs) for the PrimeKG dataset using TransE, which represents entities and relationships as embeddings with a linear translation relationship between them. Since PrimeKG includes a reversed triple $(\mathbf{o}, \mathbf{r}, \mathbf{s})$ for each original triple $(\mathbf{s}, \mathbf{r}, \mathbf{o})$, we remove the former to preserve the directed nature of the graph for training TransE. 

We utilize the implementations of the PrimeKG dataset and TransE model from PyKEEN \cite{ali2021pykeen}. The model is trained for 100 epochs with a learning rate of $10^{-3}$. Training a KGE model often involves generating negative triples to prevent under-fitting. One method for doing that is by assuming that every possible triple not present in the knowledge graph is incorrect (closed world assumption). However, as knowledge graphs are inherently incomplete, this often leads to over-fitting. Therefore, we use the stochastic local closed world assumption (sLCWA) for training, which generates a random subset of all possible negative triples $(\mathbf{s}, \mathbf{r}, \mathbf{o'}) \notin \text{PrimeKG}$ for each $(\mathbf{s}, \mathbf{r}, \mathbf{o}) \in \text{PrimeKG}$. This method strikes a balance between under-fitting and the closed world approximation, while the stochasticity helps keep computational complexity lower than full LCWA.

\subsection{Unconditional Generative Model}

Score-based models define the forward diffusion $q$ of a graph $ \textbf{G}_t = (\textbf{X}_t, \textbf{E}_t) $ with a Stochastic Differential Equation (SDE):

\begin{equation}
\label{eq:sde}
	\mathrm{d}\textbf{G}_t = \textbf{f}_t(\textbf{G}_t) + g_t\mathrm{d}\textbf{w}
\end{equation}

where $\textbf{w}$ is the standard Wiener process and $\textbf{f}_t$ and $g_t$ are the coefficients of linear drift and scalar diffusion, respectively.

The generative process is initiated with a graph $\textbf{G}_0$ and iteratively diffused to generate $\textbf{G}_T$. 

The stochastic forward process described in \eqref{eq:sde} can be used for generation by solving it's reverse-time version:

\begin{equation}
\label{eq:reversesde}
	\mathrm{d}\textbf{G}_t = [\textbf{f}_t(\textbf{G}_t)-g_t^2 \nabla_{\textbf{G}_t} \log{p_{t}(\textbf{G}_t})]\mathrm{d}\overline{t} + g_t\mathrm{d} \overline{\textbf{w}}
\end{equation}

where $\overline{t}$ and $\overline{\textbf{w}}$ represent a reverse time-step and stochastic process. A score network $s_{\theta}$ is used to approximate $\nabla_{\textbf{G}_t} \log{p_t(\textbf{G}_t})$, where $p_t(\textbf{G}_t)$ is the probability density of the diffused graph at time $t$, and simulate the reverse process in time to generate $\textbf{G}_{t-1}$.

The parameters $\boldsymbol{\theta}$ are obtained using an adaptation of the loss function by Song et al. \cite{song2021mri3}: 

\begin{eqnarray}
	\boldsymbol{\theta}^* &=& \arg\min_{\boldsymbol{\theta}} \mathbb{E}_{t} \left\{ \lambda(t) \mathbb{E}_{\textbf{G}_0} \mathbb{E}_{\textbf{G}_t|\textbf{G}_0} \left[ \left|\left| s_{\theta}(\textbf{G}_t, t) - \right. \right. \right. \right. 
   \nonumber\\
   &&\left. \left. \left. \left. \nabla_{\textbf{G}_t}\log p_t(\textbf{G}_t|\textbf{G}_0)  \right|\right|_2^2 \right] \right\}
\end{eqnarray}

Our unconditioned model is based on the GDSS architecture \cite{jo22GDSS}, trained to sample compounds from the ZINC250k dataset \cite{irwin2012zinc}. 

\subsection{Conditional Generative Model}

Guiding the stochastic process can be achieved by adding the conditioning information $\textbf{y}$ at each diffusion step, converting $p_t(\textbf{G}_t)$ from \eqref{eq:reversesde} to $p_{t}(\textbf{G}_t|\textbf{y})$. The score network is then used to approximate the modified gradient $\nabla_{\textbf{G}_t} \log{p_{t}(\textbf{G}_t}|\textbf{y})$. The conditional distribution is rearranged to give 

\begin{equation}
    \label{eq:breakdown}
	\nabla_{\textbf{G}_t} \log{p_{t}(\textbf{G}_t}|\textbf{y}) = \nabla_{\textbf{G}_t}\log p_{t}(\textbf{G}_t) + \nabla_{\textbf{G}_t}\log p_{t}(\textbf{y}|\textbf{G}_t)
\end{equation}

The term $\nabla_{\textbf{G}_t}\log p_{t}(\textbf{y}|\textbf{G}_t)$ in \eqref{eq:breakdown} steers the model towards optimizing for the condition, while the first term $\nabla_{\textbf{G}_t}\log p_{t}(\textbf{G}_t)$ introduces variation into the trajectory and helps explore newer regions. 

In order to estimate the conditional term above, we consider two noising processes $q$ and $\hat{q}$ for the unconditioned and conditioned cases, respectively. 

Using a result by Dhariwal et al. \cite{dhariwal2021diffusion}, the conditional term can be written as:

\begin{equation}
	\hat{q}(\textbf{G}_{t-1}|\textbf{G}_t, \textbf{y}) \propto  q(\textbf{G}_{t-1}|\textbf{G}_t) \hat{q}(\textbf{y}|\textbf{G}_{t-1})
\end{equation}

Using a first-order approximation for the reverse process, we obtain: 

\begin{equation}
\label{eq:prevstep}
	\log \hat{q}(\textbf{y}|\textbf{G}_{t-1}) \approx \log \hat{q}(\textbf{y}|\textbf{G}_t) + \left<\nabla_{\textbf{G}_{t}} \log \hat{q}(\textbf{y}|\textbf{G}_t), \textbf{G}_{t-1} - \textbf{G}_t\right>
\end{equation}

We now train a regressor network $P_{\phi}$ to estimate $\textbf{y}$ from a noised version of the graph $\textbf{G}_t$ and use the gradients of this network to guide the generative process. We make the assumption that the conditioned process is effectively a distribution centered at the estimated knowledge-based embedding $P_{\phi}(\textbf{G}_t)$, i.e.,

\begin{equation}
	\hat{q}(\textbf{y}|\textbf{G}_t) \sim \mathcal{N}(P_{\phi}(\textbf{G}_t), \sigma_{y}^2\mathbf{I})
\end{equation}

This leads to the overall approximation in Eq \eqref{eq:prevstep} to be: 

\begin{eqnarray}
	\log \hat{q}(\textbf{y}|\textbf{G}_{t-1}) &\approx& \left<\nabla_{\textbf{G}_t} \left|\left| \textbf{y} - P_{\phi}(\textbf{G}_t) \right|\right|_2^2, \textbf{G}_{t-1}\right> \nonumber\\ &&+ 
    \text{terms independent of } \textbf{G}_{t-1}
\end{eqnarray}

The implementation of the score-based network is based on MOOD \cite{pmlr-v202-lee23f}. The details of the regressor network $P_{\phi}$ are described below. 

\subsection{Context Regressor Network}

To guide the conditional generation process, we create a trainable network to estimate knowledge-based embeddings $\textbf{y}$ from a noised version of an input molecular graph $\textbf{G}_T$. $P_{\phi}(\textbf{G}_T)\approx\textbf{y}$ is used to implement a modified version of the classifier guidance algorithm by \cite{sohl2015deep}. While previous work by \cite{pmlr-v202-lee23f} and \cite{vignac2023digress} uses a similar algorithm to guide conditional generative processes, ours is a different approach that utilizes a stack of graph attention layers to estimate knowledge-based  embeddings, effectively creating a map between chemical space and KGE space.

 $P{_\phi}(\textbf{G}) = P_{\phi}(\textbf{X}, \textbf{E})$ is constructed by first passing the feature $\textbf{X}$ and adjacency matrices $\textbf{E}$ through an aggregation operation: $H^1 = \sigma(\textbf{E}\textbf{X}W^0_{\phi})$
 
 We then use a stack of self-attention layers:
 \begin{equation}
 	h_i^{l+1} = \sum_{j}\alpha_{ij} ~\cdot~ \mathbf{W^l}h_j^l
 \end{equation}
 where
 \begin{equation}
 	\alpha_{ij} = \text{SoftMax}(\sigma(\mathbf{W_a}^l \cdot [\mathbf{W}^l h^l_i ~\oplus~ \mathbf{W}^l h^l_j]))
 \end{equation}

Here, $\textbf{W}^l, \textbf{W}_a^l$ are learnable parameters at the $l$-th layer. The stack of attention layers produces a final output of dimension $|\textbf{y}|$.

\subsection{Interpolation in chemical space}

Upon obtaining the space of embeddings for the Knowledge Graph entities, we devise an interpolation process between the embeddings of two entities to generate new embeddings, akin to generating new molecules by interpolating between two known molecules. This process is well explored in image-based inputs \cite{pmlr-v139-oring21a}, and also in the molecular latent space generated by graph-based autoencoders \cite{du2023chemspace}. Our work extends this to the knowledge graph space, where we interpolate between the embeddings of two entities to generate new embeddings and use them to in the generative process of molecules. 

Given a set of target genes $\textbf{t}_i$, we can devise a process to generate a drug that maximizes binding to all the targets. We can formulate this as a regression problem, where we aim to predict the drug embedding $\textbf{y}$ that minimizes the distance between the sum of the drug and target gene embeddings.

\begin{equation}
 \textbf{y} = \arg\min_{\textbf{d} \in \mathcal{D}} \sum_{i} \left|\left| \textbf{d} + \textbf{r}_i - \textbf{t}_i \right|\right|_2^2
\end{equation}

Here, $r$ can potentially be the embedding for the \texttt{drug/protein} relation, and $\textbf{t}_i$ is the embedding for the target gene $i$, and $\mathcal{D}$ is the set of all possible linear combinations of known drug embeddings. 

\begin{figure*}
	\centering
	\includegraphics[width=.98\textwidth]{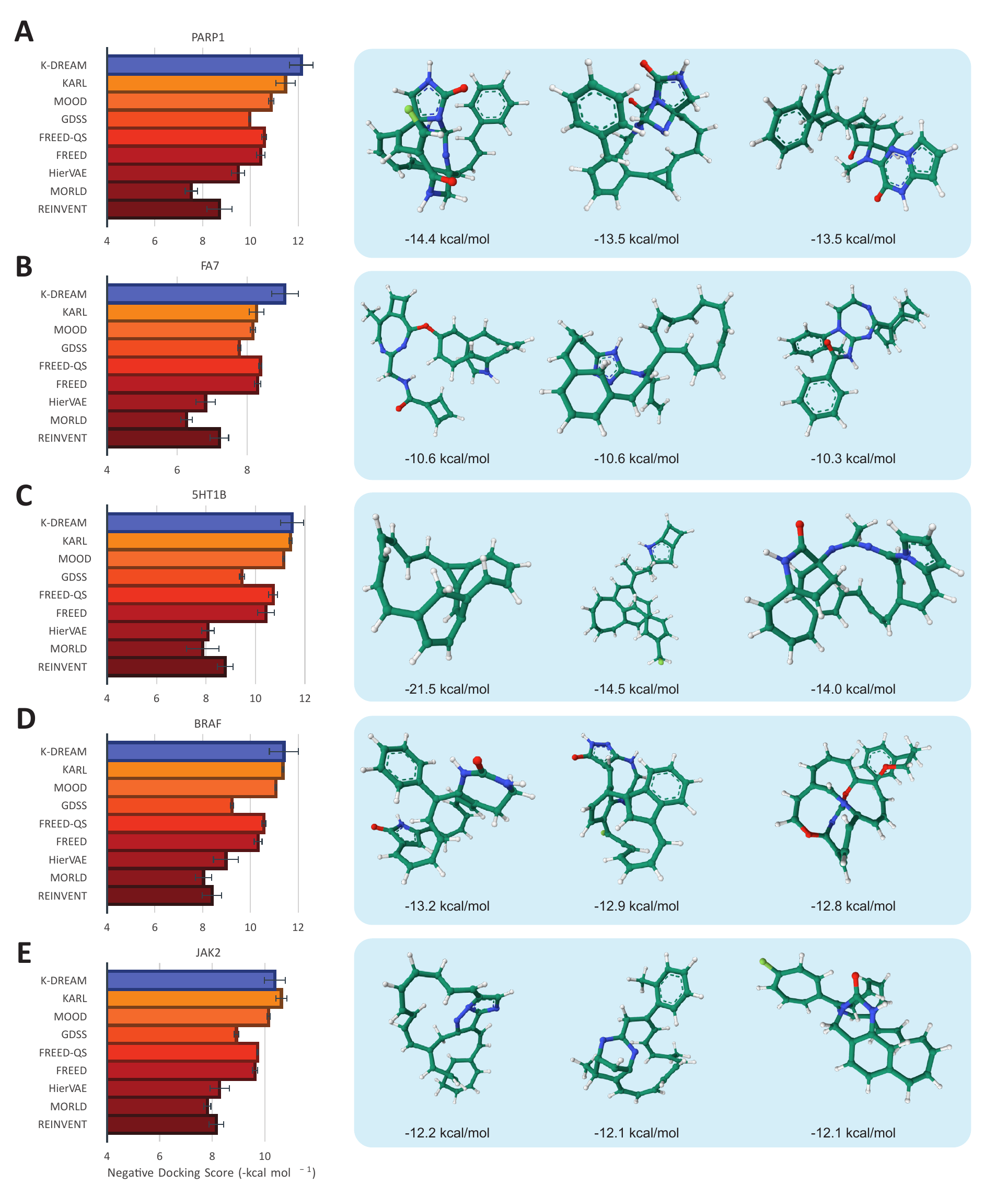}
	\caption{\rev{\textbf{Docking scores. } The mean docking scores of the top 5\% molecules generated by K-DREAM are compared against baseline models for five protein targets (left). For each protein, we also show the top 3 molecules generated by K-DREAM, along with their docking scores (right).}}
	\label{fig:dock}
\end{figure*}

\section{Results}

\begin{figure*}
	\centering
	\includegraphics[width=.98\textwidth]{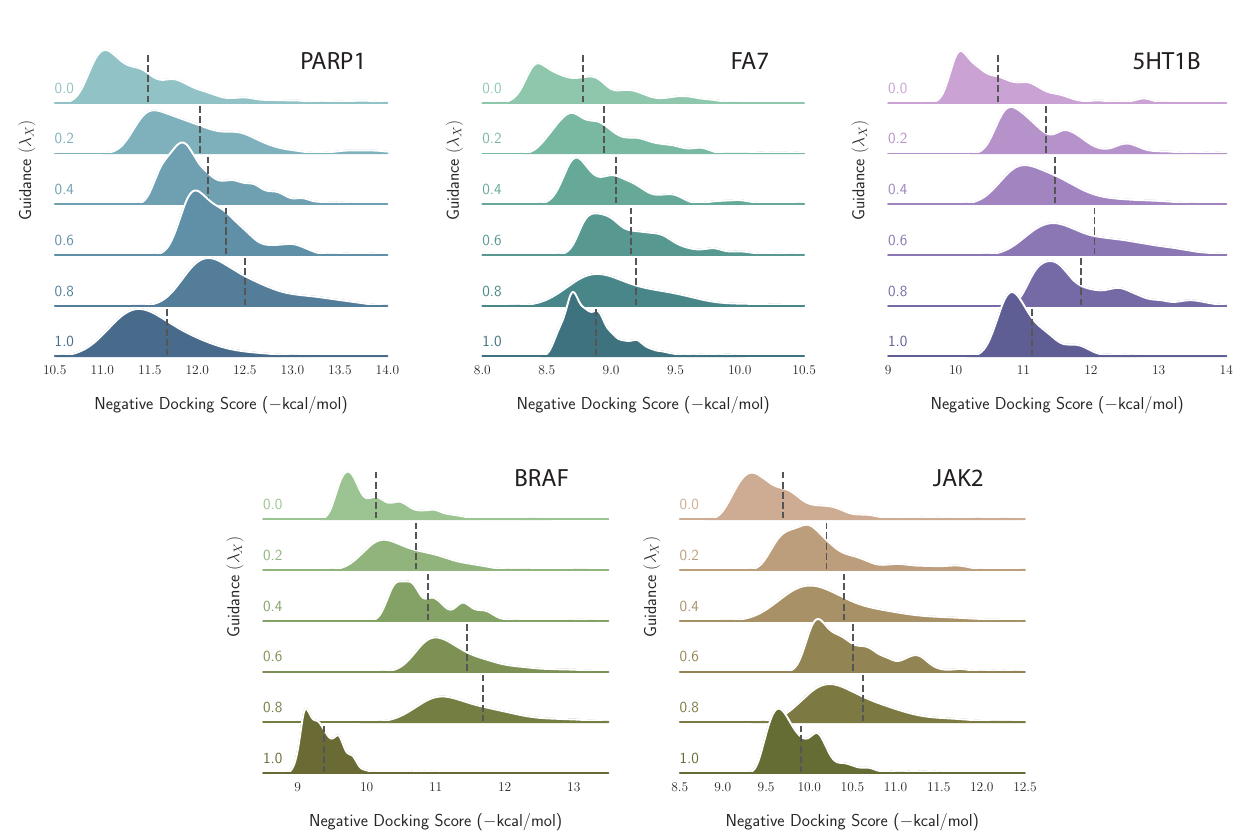}
	\caption{\textbf{Molecular docking score distribution with varying guidance levels. } The extent of guidance is controlled with a hyperparameter $\lambda_X$ that determines the weight of the guidance term in the loss function.}
	\label{fig:ridgeline}
\end{figure*}

K-DREAM builds upon recent developments in diffusion models by incorporating a regression-based guidance mechanism informed by biomedical knowledge graphs. The model employs a high-dimensional regression approach that maps molecular structures to a latent space encoded with biomedical information derived from the graph. We extend the typical formulation of classifier-based diffusion models that often rely on simpler scalar metrics \cite{pmlr-v202-lee23f} or discrete classifications for guidance. The regression network creates a mapping between chemical structures and  knowledge-derived embedding space, providing a means to guide the diffusion process based on biological context. By operating in a continuous, high-dimensional latent space, K-DREAM allows for nuanced control over the generated molecules' properties. 

We trained the Knowledge Graph Embedding (KGE) model, TransE, on the PrimeKG \cite{Chandak2023} dataset to generate knowledge graph embeddings. The dataset was preprocessed by removing duplicate entries and reversed triples to preserve the directed nature of the graph. Using the stochastic local closed world assumption (sLCWA) to generate negative triples, we balanced between underfitting and computational efficiency. The resulting embeddings were visualized using Uniform Manifold Approximation and Projection (UMAP) \cite{mcinnes2018umap}, revealing clear separation between different entity classes in the projected space. This separation demonstrates the model's ability to capture meaningful semantic relationships between entities, with clusters forming around related biological concepts such as genes, proteins, diseases, and drug compounds. We use edge bundling \cite{zhou2013edge} to visualize relations between the embeddings, enabling us to see the connections between entities more clearly, highlighting how certain entities are interlinked through various biological processes. They signify important biological interactions, such as drug-target associations, gene regulation pathways, and protein-protein interactions. 

To bridge the gap between chemical structures and knowledge graph embeddings, we developed a neural network with graph attention layers called the Context Regressor Network (CRN). Constructed using a stack of graph attention layers, this network maps molecular graphs to their corresponding knowledge-based embeddings. Crucially, the training process involved augmenting the input data by creating noised versions of each chemical structure. These noised variants were trained to map to the same embedding as the original molecule. This approach ensures that structurally similar molecules in chemical space - those within a ``neighborhood" of a given chemical structure - are mapped to the same point in the embedding space. This design choice preserves local chemical relationships while allowing for improved guidance in the generative process.

The underlying unconditioned generative model is a score-based model that samples the molecular distribution of the ZINC250k dataset, a collection of 250k commerically available chemical compounds \cite{irwin2012zinc}. The objective of the guidance mechanism is to minimize the Euclidean distance between the embedding of the molecule at the current diffusion step and the target embedding. This is modeled as a multidimensional Gaussian function centered at the target embedding. The contribution of the guidance term to the score is controlled by a hyperparameter $\lambda_X$, which determines the weight of the guidance term relative to the underlying score-based model. In our figures, we display the negated docking scores for clarity, as lower (more negative) scores indicate stronger binding affinity. 


\begin{table*}[ht!]
    \centering
    \resizebox{.98\textwidth}{!}{
    \renewcommand{\arraystretch}{0.8}
    \renewcommand{\tabcolsep}{1.5mm}
    \begin{tabular}{l@{\hskip 0.1in}ccccc}
    \toprule
        \multirow{2.5}{*}{Method} & \multicolumn{5}{c}{Target protein} \\
    \cmidrule(l{2pt}r{2pt}){2-6}
        & parp1 & fa7 & 5ht1b & braf & jak2 \\
    \midrule
        REINVENT~\cite{olivecrona2017molecular} & \phantom{0}-8.702~\scriptsize($\pm$~0.523) & -7.205~\scriptsize($\pm$~0.264) & \phantom{0}-8.770~\scriptsize($\pm$~0.316) & \phantom{0}-8.392~\scriptsize($\pm$~0.400) & \phantom{0}-8.165~\scriptsize($\pm$~0.277) \\
        MORLD~\cite{jeon20morld} & \phantom{0}-7.532~\scriptsize($\pm$~0.260) & -6.263~\scriptsize($\pm$~0.165) & \phantom{0}-7.869~\scriptsize($\pm$~0.650) & \phantom{0}-8.040~\scriptsize($\pm$~0.337) & \phantom{0}-7.816~\scriptsize($\pm$~0.133) \\
        HierVAE~\cite{jin2020hierarchical} & \phantom{0}-9.487~\scriptsize($\pm$~0.278) & -6.812~\scriptsize($\pm$~0.274) & \phantom{0}-8.081~\scriptsize($\pm$~0.252) & \phantom{0}-8.978~\scriptsize($\pm$~0.525) & \phantom{0}-8.285~\scriptsize($\pm$~0.370) \\
        FREED~\cite{yang21freed} & -10.427~\scriptsize($\pm$~0.177) & {-8.297}~\scriptsize($\pm$~0.094) & -10.425~\scriptsize($\pm$~0.331) & -10.325~\scriptsize($\pm$~0.164) & \phantom{0}-9.624~\scriptsize($\pm$~0.102) \\
        FREED-QS~\cite{yang21freed} & -10.579~\scriptsize($\pm$~0.104) & {-8.378}~\scriptsize($\pm$~0.044) & -10.714~\scriptsize($\pm$~0.183) & -10.561~\scriptsize($\pm$~0.080) & \phantom{0}-9.735~\scriptsize($\pm$~0.022) \\
        GDSS~\cite{jo22GDSS} & \phantom{0}-9.967~\scriptsize($\pm$~0.028) & -7.775~\scriptsize($\pm$~0.039) & \phantom{0}-9.459~\scriptsize($\pm$~0.101) & \phantom{0}-9.224~\scriptsize($\pm$~0.068) & \phantom{0}-8.926~\scriptsize($\pm$~0.089) \\
        MOOD~\cite{pmlr-v202-lee23f}  &
        {-10.865}~\scriptsize($\pm$~0.113) & -8.160~\scriptsize($\pm$~0.071) & {-11.145}~\scriptsize($\pm$~0.042) & {-11.063}~\scriptsize($\pm$~0.034) & {-10.147}~\scriptsize($\pm$~0.060) \\
        \rev{KARL~\cite{malusare2024improving}} & \rev{-11.475~\scriptsize($\pm$~0.410)} &\rev{-8.270~\scriptsize($\pm$~0.211)} & \rev{-11.149~\scriptsize($\pm$~0.072)} & \rev{-11.364~\scriptsize($\pm$~0.035)} & \rev{-10.636~\scriptsize($\pm$~0.212)} \\
        \midrule
        \textsc{K-DREAM} (zero guidance)  & {-10.918}~\scriptsize($\pm$~0.592) & {-8.390}~\scriptsize($\pm$~0.352) & {-9.679}~\scriptsize($\pm$~0.345) & {-10.402}~\scriptsize($\pm$~0.585) & {-10.057}~\scriptsize($\pm$~0.483) \\
        \textsc{K-DREAM}  & \textbf{-12.137}~\scriptsize($\pm$~0.490) & \textbf{-9.078}~\scriptsize($\pm$~0.381) & \textbf{-11.484}~\scriptsize($\pm$~0.459) & \textbf{-11.407}~\scriptsize($\pm$~0.612) & \textbf{-10.391}~\scriptsize($\pm$~0.0.395) \\
    \bottomrule
    \end{tabular}}
    \vspace{4pt}
        \caption{ \textbf{Top 5\% docking score (kcal/mol). } The results are the means and the standard deviations of 5 runs. Previous results sourced from \cite{pmlr-v202-lee23f}.}		
    \label{tab:baselines}
\end{table*}

\subsection{K-DREAM achieves state-of-the-art docking scores in targeted drug design tasks}
\label{sec:result1}

The evaluation of targeted drug design is conducted on five protein targets selected from the DUD-E database \cite{Mysinger2012}: PARP1, JAK2, FA7, 5HT1B, and BRAF. These proteins, chosen for their high AUROC in docking score prediction using QuickVina 2 \cite{alhossary2015fast}, represent critical targets in various therapeutic areas. PARP1, a DNA repair enzyme, is implicated in cancer therapy resistance. JAK2, a tyrosine kinase, plays a crucial role in hematopoiesis and is a target for myeloproliferative disorders. FA7, or coagulation factor VII, is essential in the blood clotting cascade and a potential target for anticoagulant therapies. 5HT1B, a serotonin receptor subtype, is involved in mood regulation and migraine pathophysiology. BRAF, a serine/threonine kinase, is frequently mutated in melanoma and other cancers, making it a prime target for oncology drug development.  

Docking scores are computed using QuickVina 2 through molecular docking simulations to provide a quantitative measure of the binding affinity between a ligand and a protein target. These scores predict the likelihood of a compound's interaction with its intended target. Lower (more negative) docking scores indicate stronger predicted binding, suggesting potentially higher efficacy of the drug candidate. 

\rev{We compare K-DREAM against several categories of molecular generative models. Variational autoencoder-based methods include HierVAE \cite{jin2020hierarchical}, which generates molecular graphs using structural motifs as building blocks through hierarchical encoding from atoms to connected motifs. Diffusion-based approaches include GDSS ~\cite{jo22GDSS}, which models joint node-edge distributions via stochastic differential equations, and MOOD \cite{pmlr-v202-lee23f}, a score-based diffusion model that incorporates out-of-distribution control and uses property prediction network gradients to guide generation toward desired chemical properties. Reinforcement learning methods include REINVENT \cite{olivecrona2017molecular}, MORLD \cite{jeon20morld}, FREED \cite{yang21freed}, and FREED-QS \cite{yang21freed}, which optimize molecular properties through reward-based learning in chemical space. We also compare against KARL \cite{malusare2024improving}, another knowledge-augmented approach, and evaluate K-DREAM with zero guidance to demonstrate the contribution of knowledge graph embeddings.}

K-DREAM generated 3000 molecules for each protein target. These molecules were subsequently ranked based on their docking scores, with the top 5\% selected for analysis. The mean and standard deviation of docking scores for this subset were calculated for each protein target (Figure \ref{fig:dock}). Following previous work, we applied filters based on Quantitative Estimate of Drug-likeness (QED) \cite{bickerton2012quantifying}, Synthetic Accessibility (SA) \cite{ertl2009estimation} score, and Tanimoto similarity \cite{Bajusz2015} to ensure the drug-likeness and synthetic feasibility of the generated molecules. The QED threshold of 0.5 ensures favorable physicochemical properties, while the SA score cutoff of 0.44 promotes synthetic feasibility. A maximum Tanimoto similarity of 0.4 to known actives maintains novelty while preserving desirable structural features. These criteria collectively ensure that K-DREAM generates not only potent but also practically viable novel drug candidates. The mean validity and uniqueness of the generated molecules is greater than 99.8\%. 

K-DREAM achieved mean docking scores of -12.13 kcal/mol for PARP1, -10.39 kcal/mol for JAK2, -9.08 kcal/mol for FA7, -11.48 kcal/mol for 5HT1B, and -11.41 kcal/mol for BRAF. We compare against baselines that use reinforcement learning \cite{yang21freed,olivecrona2017molecular,jeon20morld}, variational autoencoders \cite{jin2020hierarchical}, and score-based models \cite{jo2022score,pmlr-v202-lee23f}. K-DREAM outperforms these models across all protein targets, demonstrating its superior ability to generate molecules with high predicted binding affinity. \rev{We note the improved performance of K-DREAM relative to the KARL model, which also uses contextual information from knowledge graphs, with an explicit reward function for heuristic scores. The design of the Context Regressor Network, more robust and interpretable KGE model, and the resulting improved predicted target embeddings contribute to the gains in performance.} Additional details regarding the baseline models as well as the generation hyperparameters can be found in the Supplementary Material.

\subsection{K-DREAM interpolates in chemical space to generate multi-target compounds}

\begin{figure}
	\centering
	\resizebox{0.48\textwidth}{!}{\input{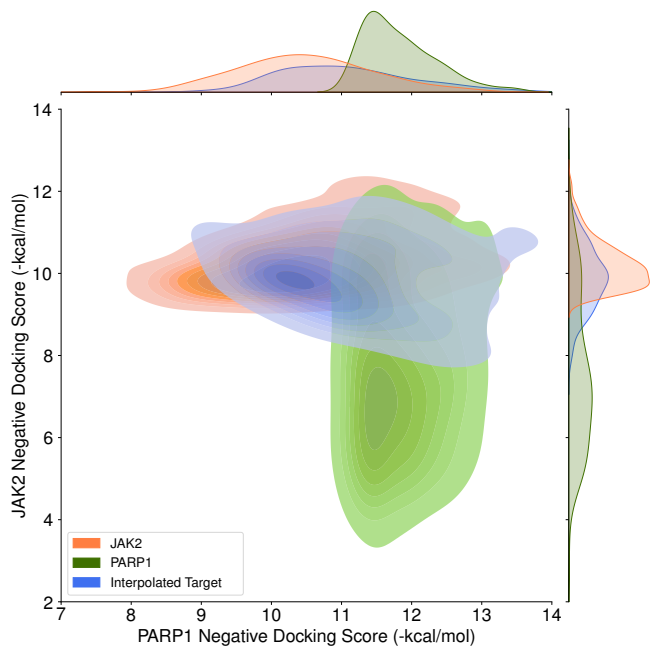}}
	\caption{\textbf{Multi-target Drug Design.} The distribution of the top 10\% of molecules (ranked by docking score) generated by K-DREAM for PARP1-targeted (green), JAK2-targeted (orange), and an interpolated target (blue), evaluated by their docking scores with PARP1 (x-axis) and JAK2 (y-axis). We can see that the interpolated molecules achieve a balance between the two target proteins with a high mean score on both axes.}
	\label{fig:expt2}
\end{figure}

Multi-target drug design presents a promising avenue for developing more effective therapeutics, particularly for complex diseases with interconnected pathways. \rev{Previous work \cite{winter} in this the field has cast this problem into the form of multi-objective latent space optimization, which uses combinations of \textit{in-silico} predictions to produce better candidates. Our work improves upon this foundational idea by leveraging the rich contextual information of the biomedical knowledge graph.} K-DREAM's ability to interpolate between target embeddings enables the generation of molecules with tailored multi-target profiles. From our set of five target proteins from the previous experiment, we select JAK2 and PARP1 for a multi-target design task. The co-activation of JAK2 and PARP1 in hepatitis virus-associated liver cancer underscores the need for multi-target therapeutic approaches \cite{Cherng2021}. JAK2 is upregulated in virus-associated hepatocellular carcinoma (vHCC) and contributes to tumor progression. Concurrently, PARP1, a DNA repair enzyme, promotes therapy resistance in various cancers, including HCC \cite{yu2022parg}. Multi-target drugs addressing both JAK2 and PARP1 could potentially offer a more streamlined approach to combat the complex molecular landscape of HCC while mitigating the challenges associated with combination therapies.

\subsection{Ablation Studies}

Table \ref{tab:baselines} compares the performance of K-DREAM without the guidance information from the knowledge graph for all five protein targets. We observe a significant decrease in docking scores across all targets, demonstrating the effectiveness of the embeddings in the guidance process 

For quantitative evaluation of multi-target interpolation, we generated three compound sets consisting of 3000 molecules each: PARP1-targeted ($P$), JAK2-targeted ($J$), and interpolated-target ($I$) (Figure \ref{fig:expt2}). PARP1-targeted compounds scored ($\mu_{P,P} = -11.84$, $\sigma = 0.54$) when evaluated against PARP1 and ($\mu_{P,J} = -7.53$, $\sigma = 1.84$) against JAK2. JAK2-targeted compounds achieved ($\mu_{J,P} =- 10.48$, $\sigma = 0.95$) and ($\mu_{J,J} = -10.21$, $\sigma = 0.57$) respectively. Interpolated compounds attained ($\mu_{I,P} = -11.06$, $\sigma = 0.94$) and ($\mu_{I,J} = -9.89$, $\sigma = 0.79$). Post-hoc Hotelling's $T^2$ tests confirmed significant distinctions between all pairs ($P$ vs $I$: $T^2 = 354.22$, $F = 176.67$; $J$ vs $I$: $T^2 = 65.52$, $F = 32.68$; $P$ vs $J$: $T^2 = 935.85$, $F = 466.95$; all $p < 0.001$). Mahalanobis distances between centroids ($D_{P,I} = 2.49$, $D_{J,I} = 0.67$, $D_{P,J} = 3.01$) demonstrated the interpolated set's intermediate position. Interpolated compounds' PARP1 and JAK2 scores showed a slight negative correlation ($r = -0.16$, $p < 0.05$). The proportion of compounds exceeding predefined thresholds (PARP1 $\leq -11.5$, JAK2 $\leq -10.0$) was highest for interpolated compounds ($15.4\%$) compared to PARP1-targeted ($9.6\%$) and JAK2-targeted ($12.5\%$) sets, validating K-DREAM's ability to generate compounds with balanced dual-inhibition profiles. (All docking scores are stated in kcal/mol)

The Supplementary Material contains additional ablation studies with reference to the embedding dimension.

\section{Discussion}\label{sec:discussion}

K-DREAM achieved state-of-the-art performance in targeted drug design, surpassing baseline models in generating molecules with superior docking scores. Unlike baseline models \rev{like GDSS, MOOD and others} that directly incorporate docking score information, K-DREAM leverages knowledge graph representations to guide molecule generation. This approach demonstrates that biomedical knowledge graphs contain sufficient information to steer the generative process towards biologically relevant molecules. The model's ability to generate high-scoring molecules without explicit docking information suggests it captures complex relationships between molecular structures and biological targets, potentially leading to more diverse and innovative drug candidates.

\rev{While docking scores provide standardized computational metrics, they have significant limitations as primary evaluation measures, relying on static protein structures and simplified scoring functions that cannot capture dynamic conformational changes or cellular complexity. Docking scores also fail to predict pharmacokinetic properties, selectivity, or functional activity essential for therapeutic efficacy. Future validation of K-DREAM requires experimental studies including biochemical binding assays, cell-based functional activity tests, and \textit{in vivo} pharmacokinetic studies to confirm that knowledge graph guidance translates to genuine improvements in biological activity and therapeutic potential beyond computational predictions.}

In creating multi-target compounds, the existence of the IFNGR-JAK-STAT-PARP1 pathway between our chosen targets PARP1 and JAK2 in the knowledge graph exemplifies the interconnected nature of biological systems, which traditional drug discovery methods often struggle to address comprehensively. K-DREAM's integration of knowledge graphs with generative models represents a significant advancement in capturing and leveraging these complex relationships. By encoding pathway information and protein interactions in the knowledge graph, K-DREAM can generate compounds that simultaneously modulate multiple targets within a relevant biological context. Statistical analysis validates this capability: interpolated compounds achieve balanced binding to both targets ($\mu_{I,P} = -11.06$, $\mu_{I,J} =- 9.89$) with only minimal trade-offs between PARP1 and JAK2 binding ($r = -0.16$, $p < 0.05$). Most importantly, the interpolated compounds show superior dual-targeting with $15.4\%$ exceeding both binding thresholds compared to $9.6\%$ and $12.5\%$ for single-target optimization. This approach enables the exploration of chemical space guided by system-level understanding, potentially yielding molecules with optimized efficacy and reduced off-target effects. The ability to interpolate between target embeddings further allows for fine-tuning of multi-target profiles, offering a powerful tool for rational drug design in complex diseases where multiple pathways contribute to pathogenesis.

A key dependence of the model is in the weight of the guidance mechanism relative to the unconditioned generative process, controlled by the hyperparameter $\lambda_X$, which quantifies the contribution of the multidimensional Gaussian distribution centered at the target embedding. Varying $\lambda_X$ has the effect of changing the variance of the Gaussian, with higher $\lambda_X$ values corresponding to a sharper peak around the target, leading to improved docking scores. We observe an optimal value for $\lambda_X$, since an increasingly sharper peak might lead to sub-optimal guidance in the initial stages of the diffusion process (Figure \ref{fig:ridgeline}). 

We note that the increased standard deviation in K-DREAM's scores, despite higher mean scores, could be attributed to the curse of dimensionality in high-dimensional Gaussian distributions \cite{berisha2021digital}. This becomes a factor due to our guidance mechanism, which uses a Gaussian distribution to guide the generative process towards target embeddings. In high-dimensional spaces, the probability mass of a multivariate Gaussian concentrates in a thin shell away from the mean. This concentration effect leads to a larger variety of molecules, as the generative process is guided towards points on this shell rather than converging to a single central point. The volume of this shell grows exponentially with dimensionality, providing a vast space for diverse, high-scoring candidates. Consequently, while K-DREAM consistently produces molecules with improved docking scores, the diversity within this high-scoring set contributes to larger standard deviations. This behavior aligns with the model's objective of exploring a broader, biologically relevant chemical space, rather than converging to a narrow set of solutions.

\section{Conclusion}

This study demonstrates that biomedical knowledge graphs contain valuable information that can effectively guide molecular generation toward biologically relevant compounds. By embedding this knowledge into a diffusion-based generative framework, K-DREAM produces drug candidates with improved binding potential across multiple therapeutic targets while maintaining chemical validity and diversity.
The framework's performance in targeted drug design tasks consistently exceeded baseline models across all tested protein targets. Unlike previous approaches that directly optimize for docking scores, K-DREAM leverages the rich contextual information within knowledge graphs to steer the generative process toward promising regions of chemical space without explicitly incorporating target-specific scoring functions.
Perhaps most significantly, K-DREAM's ability to interpolate between target embeddings enables the generation of molecules with balanced multi-target profiles. This capability addresses a critical need in modern drug discovery, where many diseases involve multiple interconnected pathways that cannot be effectively addressed by single-target approaches.
The integration of knowledge graph embeddings with generative models creates a unified space where chemical structures and biological interactions can be jointly explored. This paradigm shift moves beyond traditional property-based optimization toward a more comprehensive, biologically informed approach to drug design.

Future research could further enhance this framework by incorporating additional biomedical knowledge sources, exploring alternative embedding techniques, and extending the methodology to address other drug design challenges such as optimizing pharmacokinetic properties alongside target binding. By continuing to bridge the gap between computational generation and biological understanding, knowledge-enhanced generative models may significantly accelerate the discovery of novel therapeutics for unmet medical needs.

\section*{Acknowledgements}

The authors acknowledge the use of ChatGPT and Claude to refine human-written text in this manuscript. 

\bibliography{exported,manual,nmi}
\bibliographystyle{ieeetr}

\onecolumn
\newpage
\appendix
\section*{Supplementary Material}
\section{Experimental Details}

\subsection{Source code}

\rev{The implementation of K-DREAM is available at this link: \url{https://github.itap.purdue.edu/Clan-labs/k-dream}}

\subsection{Knowledge Graph: Preparation and Potential Biases}

We use PrimeKG (\href{https://github.com/mims-harvard/PrimeKG}{github.com/mims-harvard/PrimeKG}), an extensive biomedical knowledge graph describing more than 4 million relationships among 129,375 entities including 17,080 diseases and 7,957 drugs. The graph consists of triples $\left<h, r, t\right>$ denoting a directed relationship $r$ from the head $h$ to the tail $t$, e.g. $\left<\texttt{drug}, \texttt{binds\textunderscore{to}}, \texttt{protein}\right>$. We pre-process the knowledge graph triples by first removing the reversed triples $\left<t, r, h\right>$ to reflect the directed nature of the relationships, and then remove duplicate triples. We obtain the SMILES ID of each drug in this dataset using the DrugBank ID, which is further used to map to the 2D graph representation of the drug molecule using RDKit \href{www.rdkit.org}{(rdkit.org)}.

\rev{Relying solely on PrimeKG introduces several potential biases in K-DREAM's molecular generation process: (i) completeness bias may arise from PrimeKG's inherent incompleteness, as knowledge graphs capture only a fraction of all biological relationships, potentially missing emerging therapeutic targets or novel interaction mechanisms, (ii) representation bias could occur due to uneven coverage across biological domains, where well-studied proteins have richer representations than understudied targets, steering generation toward better-characterized systems, (iii) curation bias, which reflects historical research priorities embedded in PrimeKG's source databases, potentially favoring certain disease areas or molecular classes, and, (iv) temporal bias, meaning recent discoveries may be underrepresented in the static embeddings. }

\section{Additional Results}
\subsection{Experiment 1: Targeted Drug Design}\

\begin{figure}
	\centering
	\includegraphics[height=0.9\textheight]{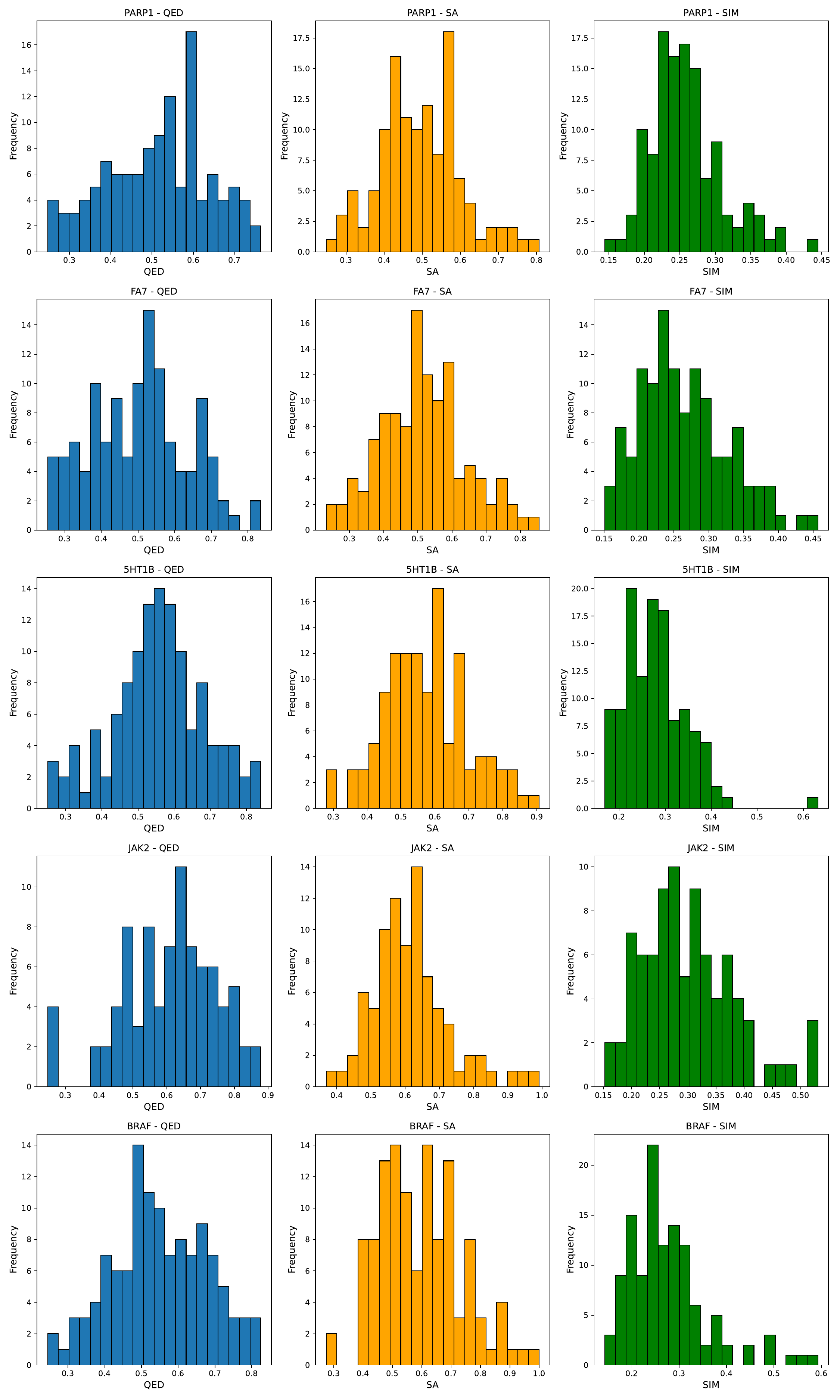}
	\caption{\textbf{Targeted Drug Design - QED, SA and SIM scores.} The distributions of Quantitative Estimate of Drug-likeness (QED), Synthetic Accessibility (SA) and Tanimoto similarity (SIM) scores for the top 5\% candidates chosen for evaluation in the targeted drug design experiment. }
	\label{fig:additionaldist}
\end{figure}

Figure \ref{fig:additionaldist} shows the distributions of Quantitative Estimate of Drug-likeness (QED), Synthetic Accessibility (SA) and Tanimoto similarity (SIM) scores for the top 5\% candidates chosen for evaluation in the targeted drug design experiment.

Table \ref{tab:baselines} provides a comparison of K-DREAM with baseline models in terms of mean docking scores for the top 5\% molecules generated for each protein target.

Table \ref{tab:hyperparameters} lists the hyperparameters used in the targeted drug design experiment. These hyperparameters are chosen by doing a systematic sweep of the guidance ($\lambda_X$) with from 0 to 1 with a step size of $0.1$.

\section{Ablation Studies}

Table \ref{tab:baselines} provides a comparison of K-DREAM with the guidance hyperparameter set to 0, in order to evaluate the impact of the guidance mechanism on the unconditioned model's performance. Figure \ref{fig:ridgeline} also shows the distribution of docking scores with varying levels of guidance.

\begin{figure}
	\centering
	\includegraphics[width=.98\textwidth]{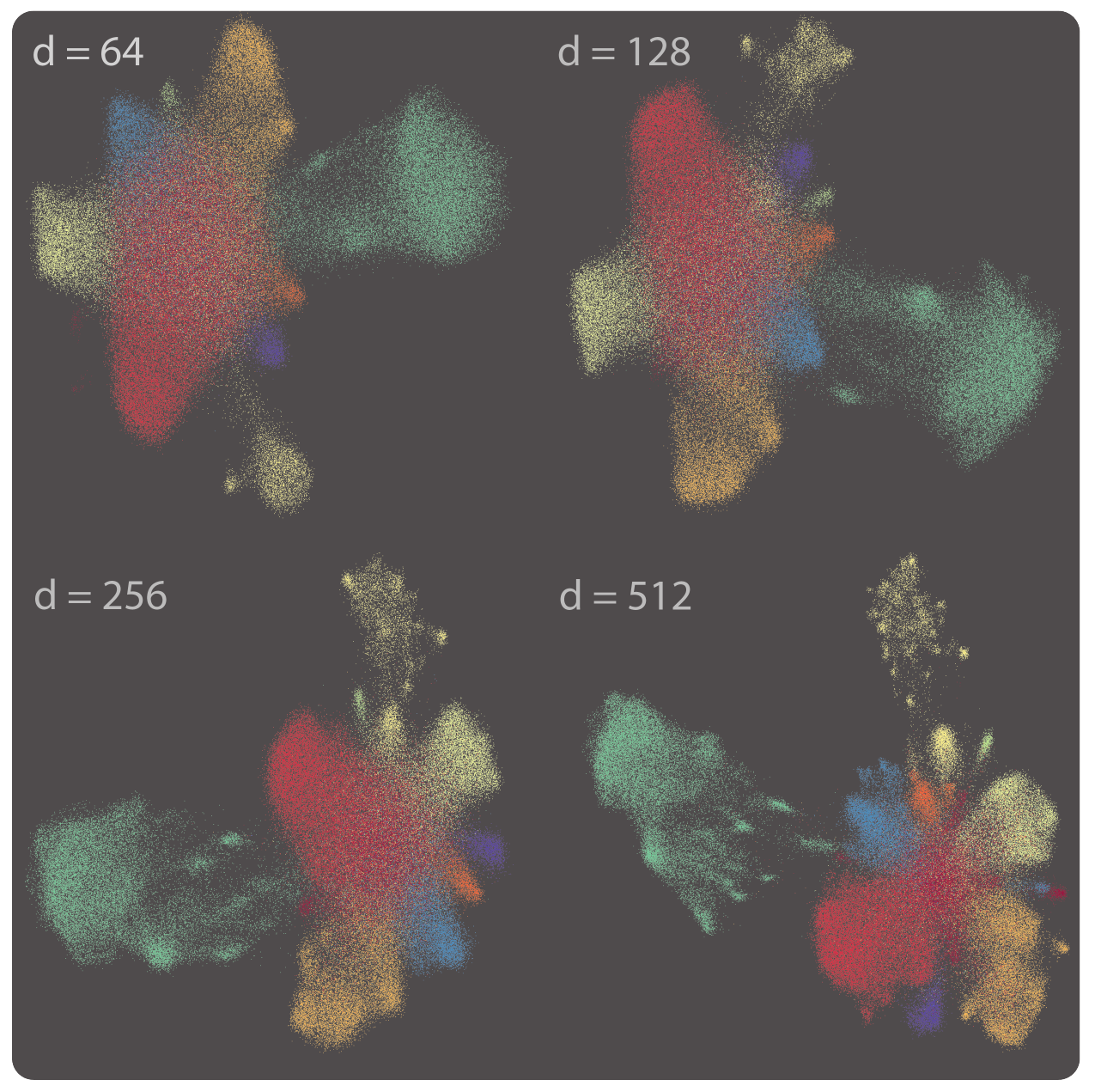}
	\caption{\textbf{Ablation Study - Dimensionality.} UMAP projections of the TransE knowledge graph embeddings with varying dimensions.}
	\label{fig:dimensionablation}
\end{figure}

Figure \ref{fig:dimensionablation} compares the clustering in UMAP projections of the TransE knowledge graph embeddings with $d=\{64, 128, 256, 512\}$. This visualization demonstrates the impact of dimensionality on the quality of the embeddings, with higher dimensions leading to better separation between entity classes.

\begin{table}[htbp]
\centering
\begin{tabular}{lccccc}
\toprule
\textbf{Metric} & \textbf{PARP1} & \textbf{FA7} & \textbf{5HT1B} & \textbf{BRAF} & \textbf{JAK2} \\
\midrule
Sampling Time (sec) & 1402 & 1438 & 1396 & 1403 & 1400 \\
Guidance ($\mathbf{X}$) & 0.5 & 0.6 & 0.7 & 0.8 & 0.95 \\
OOD & 0.03 & 0.03 & 0.03 & 0.01 & 0.00 \\
Validity & 1.0 & 1.0 & 1.0 & 1.0 & 1.0 \\
Uniqueness & 1.0 & 0.9933 & 1.0 & 0.9833 & 0.9867 \\
Novelty (sim. $<$ 0.4) & 0.8167 & 0.9067 & 0.9033 & 0.7967 & 0.8533 \\
\begin{tabular}{@{}l@{}}Novel Top 5\% DS \\ (kcal/mol) \end{tabular}
& $-12.14 \pm 0.49$ & $-9.08 \pm 0.38$ & $-11.48 \pm 0.46$ & $-11.41 \pm 0.61$ & $-10.39 \pm 0.40$ \\
Docking Time (sec) & 2654 & 2658 & 2681 & 2655 & 2655\\
\bottomrule
\end{tabular}
\caption{Hyperparameters}
\label{tab:hyperparameters}
\end{table}

\

\newpage
\section{Algorithm pseudocode}
\rev{
\begin{algorithm}
\caption{K-DREAM: Knowledge-Driven Molecular Generation}
\begin{algorithmic}[1]
\REQUIRE Target protein identifier $p$, guidance strength $\lambda$
\ENSURE Generated molecule $G_0$ with high binding affinity to $p$

\STATE \textbf{Phase 1: Knowledge Graph Embedding}
\STATE Load PrimeKG biomedical knowledge graph
\STATE Remove reversed and duplicate triples
\STATE Initialize TransE model with embedding dimension $d = 512$
\FOR{$epoch = 1$ to $100$}
    \FOR{each triple $(s, r, o)$ in knowledge graph}
        \STATE Sample negative examples using stochastic local closed world assumption
        \STATE Compute loss $\mathcal{L}$
        \STATE Update embeddings via gradient descent with learning rate $10^{-3}$
    \ENDFOR
\ENDFOR
\STATE Store entity embeddings $\{e_i\}$ and relation embeddings $\{r_j\}$

\STATE
\STATE \textbf{Phase 2: Train Context Regressor Network}
\STATE Initialize graph attention network $P_\phi$ with $L$ layers
\FOR{each drug molecule $M$ with known embedding $y$}
    \FOR{each diffusion timestep $t$}
        \STATE Generate noised version $M_t$ by adding noise to $M$
        \STATE $H^1 = \sigma(E \cdot X \cdot W^0)$ where $X$ is node features, $E$ is adjacency
        \FOR{layer $l = 1$ to $L$}
            \FOR{each node $i$}
                \STATE Compute attention: $\alpha_{ij} = \text{softmax}(\sigma(W_a^l \cdot [W^l h_i^l \oplus W^l h_j^l]))$
                \STATE Update: $h_i^{l+1} = \sum_j \alpha_{ij} \cdot W^l h_j^l$
            \ENDFOR
        \ENDFOR
        \STATE $\hat{y} = P_\phi(M_t)$ from final layer output
        \STATE Compute loss $\mathcal{L} = \|\hat{y} - y\|^2$
        \STATE Update network parameters $\phi$
    \ENDFOR
\ENDFOR

\STATE
\STATE \textbf{Phase 3: Conditional Molecular Generation}
\STATE Retrieve target embedding $y_p = e_p$ for protein $p$
\STATE Initialize random molecular graph $G_T$ at time $T$
\FOR{$t = T$ down to $1$}
    \STATE Compute unconditional score $s_\theta(G_t, t)$ from base diffusion model
    \STATE Predict embedding $\hat{y}_t = P_\phi(G_t)$ from current graph
    \STATE Compute guidance gradient $\nabla_{G_t} \|y_p - \hat{y}_t\|^2$
    \STATE Combine scores: $s_{\text{total}} = s_\theta(G_t, t) + \lambda \cdot \nabla_{G_t} \|y_p - \hat{y}_t\|^2$
    \STATE Update graph via reverse SDE: $G_{t-1} = \text{reverse\_step}(G_t, s_{\text{total}})$
\ENDFOR
\end{algorithmic}
\end{algorithm}
}

\end{document}